\documentclass[twocolumn]{webofc}

\usepackage[varg]{txfonts}   
\usepackage{hyperref}
\usepackage{url}
\hypersetup{colorlinks=true,citecolor=blue,urlcolor=blue,linkcolor=blue}
\usepackage{amsmath}
\usepackage{amssymb}
\usepackage{commath}
\usepackage[ruled,norelsize,vlined,linesnumbered]{algorithm2e}
\makeatletter
\newcommand{\removelatexerror}{\let\@latex@error\@gobble}

\makeatother
\graphicspath{ {Figures/} }
\begin{document}
\title{Path Planning for a UAV Swarm Using Formation Teaching-Learning-Based Optimization}
%
%

\author{\firstname{Van Truong} \lastname{Hoang}\inst{1}\fnsep\thanks{\email{vantruong.hoang@alumni.uts.edu.au}} \and
        \firstname{Manh Duong} \lastname{Phung}\inst{2}\fnsep\thanks{\email{duong.phung@fulbright.edu.vn}} 
}

\institute{Faculty of Missile and Gunship, Naval Academy, Nha Trang, Khanh Hoa, Vietnam 
\and
           Undergraduate Faculty, Fulbright University, Ho Chi Minh City, Vietnam 
          }

\abstract{This work addresses the path planning problem for a group of unmanned aerial vehicles (UAVs) to maintain a desired formation during operation. Our approach formulates the problem as an optimization task by defining a set of fitness functions that not only ensure the formation but also include constraints for optimal and safe UAV operation. To optimize the fitness function and obtain a suboptimal path, we employ the teaching-learning-based optimization algorithm and then further enhance it with mechanisms such as mutation, elite strategy, and multi-subject combination. A number of simulations and experiments have been conducted to evaluate the proposed method. The results demonstrate that the algorithm successfully generates valid paths for the UAVs to fly in a triangular formation for an inspection task.
}
\maketitle
\section{Introduction}\label{intro}

The demand for the coordination of multiple unmanned aerial vehicles (UAVs) is becoming increasingly common due to their ability to perform complex tasks that a single UAV cannot handle. When operating in swarms, UAVs can share information and coordinate actions to execute tasks more efficiently. Utilizing multiple UAVs also enhances system reliability, as it reduces dependence on a single UAV \cite{yang2023uav}.

When carrying out tasks, UAVs often fly in a specific formation to ensure the stability of the collected data and effective collaboration. To accomplish this, flight paths are pre-planned for the UAVs and then adjusted during operation. In the literature, a number of studies have been conducted to address the path planning problem. Several widely used algorithms include the Fast Marching Method, Bi-level, Rapidly Exploring Random Tree (RRT), and artificial potential field  \cite{lopez2021, d2019bi, huang2020, pan2021}. These methods can generate feasible flight paths for UAVs and avoid collisions with obstacles. The optimality of the generated paths, however, is not properly addressed.

Recently, nature-inspired optimization algorithms have been increasingly used for UAV path planning due to their capability to generate optimal paths at a reasonable computational complexity. Constraints on the UAV position can also be added to enable formation fly. Popular methods include the Genetic Algorithm (GA), Ant Colony (ACO), and Particle Swarm (PSO). GA is effective in solving mixed path-planning problems \cite{roberge2018}. ACO is capable of handling larger-scale path planning problems \cite{song2022}. PSO and its variants have a faster convergence rate \cite{hoang2018iros}. However, these methods have limitations when applying for UAVs. For example, GA requires intensive computation and tends to converge prematurely, ACO has a relatively slow convergence rate, and PSO may get trapped in local optima rather than finding global solutions \cite{ait2022uav}. 

Among nature-inspired optimization algorithms, the Teaching Learning Based Optimization (TLBO) \cite{rao2011tlbo} is commonly used to solve complex problems with fast convergence \cite{majumder2021, sabiha2022}. In our previous work \cite{hoang2021, hoang2023}, we employed TLBO to solve the path planning problem for a single UAV. The algorithm was able to identify reference paths that satisfied safety, dynamic, and task-specific requirements within a multi-constraint fitness function. The results showcased the potential of TLBO, requiring minimal parameter adjustments and achieving solutions within a reasonable execution time.

This work focuses on coordinating multiple UAVs for data collection. Building on our previous research on UAV path planning \cite{hoang2021}, we extend it from a single UAV to multiple UAV formation with the development of an algorithm named formation teaching-learning based optimization (FTLBO). Our approach starts with the development of a formation model, followed by the definition of a multi-objective fitness function to transform the formation problem into an optimization problem. The fitness function is designed to address three primary objectives: obtaining the shortest path, maintaining safe operation, and ensuring task efficiency. We then apply the TLBO to determine the optimal solution of the fitness function which corresponds to the path of the formation that meets the objectives. This path is used as a basis to calculate individual routes for each UAV so that they together create the desired formation. Experiments and comparisons have been conducted to evaluate the approach. The results show that our method is both practical and effective in generating paths for the UAVs to flight  when conducting a surveying task.

The structure of the paper is as follows: Section \ref{statement} introduces the formation model and the multi-objective path planning problem. Section \ref{FTLBO} details the proposed algorithm. Section \ref{secImplementation} discusses the implementation of the path planning algorithm for the UAV formation and individual UAVs. Simulation and experimental results are presented in Section \ref{results}, followed by the conclusion and future directions at the end of the paper.



\begin{figure}
	\centering
	\vspace{-3mm}
	\includegraphics[width=7cm]{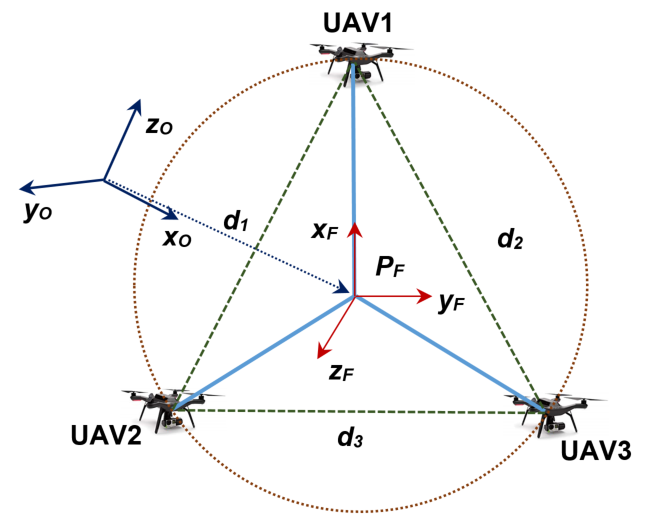}
	\caption{Inertial and formation frames in UAV formation}
	\label{form_fig}
	\vspace{-3mm}
\end{figure}

\section{Problem formulation} \label{statement}
\subsection{Formation model} \label{model}
In this work, the goal is to plan the paths for each UAV so that the UAVs will form the desired shape when flying along those paths. The rule for the formation is that each UAV will maintain the same distance to its group's centroid and to its adjacent neighbors. Figure \ref{form_fig} illustrates the formation for the case of three UAVs. Let $\{x_F, y_F, z_F\}$ be the formation frame attached to the centroid $P_F$ of the UAVs' group and $\{x_0, y_0, z_0\}$ be the inertial frame attached to a fixed point. $P_F$ is determined via position $P_n = \{x_n, y_n,z_n\}$ of UAV$_n$ as follows: 

\begin{equation}
	P_F = \dfrac{1}{N} \sum\limits_{n=1}^N P_n,
\end{equation}
where $N$ is the number of UAVs. Denote $d_n$ as the distance from UAV$_n$ to the centroid, $d_n = \norm{P_n - P_F}$, and $r_F$ as the pre-defined radius of the formation. The rule for maintaining equal distance from each UAV to the centroid can be expressed as:

\begin{equation} \label{eq2}
	d_n = r_F, \forall d_n \in \{d_1,...,d_M \}.
\end{equation}

Let UAV$_m$ and UAV$_l$ be the adjacent neighbors of UAV$_n$. The rule for maintaining equal distance between the UAVs is defined as follows:

\begin{equation} \label{eq3}
	\norm{P_n - P_m} = \norm{P_n - P_l}, \forall n \in \{1,...,M \}
\end{equation}



To compute the path for UAV$_n$, its position is determined as:

\begin{equation} \label{eq4}
	P_n = P_F + \Delta P_n,
\end{equation}
where $\Delta P_n = [e^n_x, e^n_y, e^n_z]^T$ is the adjustment to move UAV$_n$ away from the group's centroid. By substituting (\ref{eq4}) into (\ref{eq2}) and (\ref{eq3}) and solving the resulting equations, we obtain the value of $\Delta P_n$. Using this, the position of each UAV in the formation can be determined for each corresponding value of $P_F$. The formation problem is thus transformed into a path planning problem, where the objective is to generate a path for the formation's centroid that satisfies the requirements for completing the UAV's task. We address this via the definition of a multi-objective fitness function described in the next section.

\subsection{Multi-objective fitness function}
When planning a path for multiple UAVs in a group, several constraints must be met to ensure the maneuverability of each UAV, allocate sufficient operating space, and avoid obstacles. These constraints typically consist of two key aspects: path length and operational safety. In a 3D environment, additional considerations are required, such as limitations on flight altitude, turning angles, and path smoothness. In this work, constraints are integrated into the path planning process via the definition of fitness functions of the form:

\begin{equation} \label{cost_function}
\coprod {r(q)}  = \alpha\coprod\nolimits_{length} {r(q)}  + \beta\coprod\nolimits_{safe} {r(q)}  + \gamma\coprod\nolimits_{task} {r(q)},
\end{equation}
where  $\coprod r(q)$ represents the overall fitness function used to generate path $q$ from the start position $P_s = (x_s, y_s, z_s)$ to the goal position $P_f = (x_f, y_f, z_f)$;  $\coprod\nolimits_{length} {r(q)},  \coprod\nolimits_{safe} {r(q)}$, and $\coprod\nolimits_{task} {r(q)}$ are the fitness functions defined to represent the path length, collision violation and task performance, respectively; $\alpha$, $\beta$, and $\gamma$ are weight coefficients. 

The cost associated with the path length $\coprod\nolimits_{length} {r(q)}$ is determined as the sum of path segments. Specifically, we represent a flight path $q$ as a set of $m$ nodes $W_j$, also known as waypoints:
\begin{equation} 
q = \{W_1,W_2,\cdots,W_m\}.
\end{equation} 
The cost $\coprod\nolimits_{length} {r(q)}$ is then calculated as:
\begin{equation}
\coprod\nolimits_{length} {r(q)} = \sum \limits_{j=0}^{m+1} \norm{\overrightarrow{(W_j,W_{j+1}}}.
\end{equation}

To form $\coprod\nolimits_{safe} {r(q)}$ for obstacle avoidance, we represent each obstacle $k$ by a cylinder having center $O_k$ and radius $R_k$. The distance from a path segment to the cylinder is then can be used to describe the violation cost. Specifically, the violation cost between path segment $j$th and obstacle $k$ is computed as follows:

\begin{align}
	V_{j,k}(r(q)) = \begin{cases} R_k/d_{j,k} &\text{if } d_{j,k} > R_k \\
		\infty &\text{if } d_{j,k} \leq R_k,
	\end{cases}
\end{align}
where $d_{j,k}$ is the distance between the center of obstacle $k$ and path segment $j$. Summing over all $m$ segments and $K$ obstacles gives the total violation cost as follows:
\begin{equation}\label{eqViolation}
\coprod\nolimits_{safe} {r(q)} =  \frac{1}{mK} \sum \limits_{j=1}^{m}  \sum \limits_{k=1}^K V_{j,k}(r(q)).
\end{equation}

When performing tasks, UAVs are expected to fly within a specific altitude range to optimize data collection and energy efficiency. Let $h_{\text{min}}$ and $h_{\text{max}}$ represent the minimum and maximum allowable altitudes, respectively, and $h^i_j$ be the altitude of UAV$_i$ at viewpoint $j$. The cost related to the flight altitude is calculated as follows:
\begin{align}
\coprod\nolimits_{task,j} {r(q)} = \begin{cases}\nonumber
			0,  &\text{if } h_{\text{min}} \leq h^i_j \leq h_{\text{max}}\\
			(h_{\text{min}} - h^i_j),  &\text{if } h^i_j < h_{\text{min}}\\
			(h^i_j - h_{\text{max}}),  &\text{if } h^i_j > h_{\text{max}}\\
			\infty,  &\text{if } h^i_j \leq 0.
			\end{cases}
\end{align}
Denote $J$ as the total number of viewpoints that the UAVs have to collect data at. The task performance function for all UAVs is expressed as follows:
\begin{equation}\label{eqTask}
\coprod\nolimits_{task} {r(q)} =  \sum \limits_{n=1}^{N} \sum \limits_{j=1}^{J} \coprod\nolimits_{task,j} {r(q)}.
\end{equation}

\section{Formation path planning with the teaching-learning-based optimization} \label{FTLBO}

The fitness function defined in (\ref{cost_function}) converts the path planning task into an optimization problem, where the goal is to find $q$ that minimizes $\coprod\nolimits {r(q)}$. Given the complexity of $\coprod\nolimits {r(q)}$, solving it to find the exact solution is impractical. Therefore, we propose using metaheuristic optimization algorithms to obtain a sub-optimal solution within a finite time. In this work, we apply the teaching-learning-based optimization method, detailed as follows.

\subsection{Teaching-learning-based optimization (TLBO)}
The TLBO method is inspired by the learning process in a classroom, where each student represents a candidate solution. The objective is to find a process that trains students to achieve higher grades corresponding to better solutions. This process includes into two phases, teaching and learning.

In the teaching phase, each student learns from the teacher to improve their knowledge. This process is formulated by defining $T$ as the teacher's knowledge, $S_i$ as the knowledge of student $i$, and $A$ as the class average. Since $T$ corresponds to the best solution, the improvement in each student's knowledge is represented as follows:

\begin{equation}\label{eqTLBOteacher}
S^{new}_i =  S^{old}_{i} + w_0(T - \lambda A),
\end{equation}
where $w_0$ is a random number within the range $[0, 1]$, and $\lambda$ is a coefficient that takes a value of either 1 or 2 with equal probability.  To guarantee improvement, the new solution $S^{new}_i$ is carried forward to the next computation round only if it is better than the previous solution $S^{old}_{i}$.

In the learning phase, students improve their knowledge through interactions with other students. This process is represented by the following formula: 

\begin{equation}\label{eqTLBOlearner}
	S^{new}_i =  S^{old}_i + w_i \vert S_m - S_n \vert,
\end{equation}
where $m \neq n \neq i$. Similar to the teaching phase, $S^{new}_i$ is only carried forward to the next computation round if it is better than $S^{old}_{i}$. 

\subsection{TLBO enhancements for formation path planning}

When applying to formation path planning, the original TLBO can be trapped at local minima due to the complexity of the 3D environment and high-dimensional variables. To address this, we implement two enhancements including a mutation mechanism and an elite selection process as in \cite{niu2014}. The mutation mechanism is used to improve a solution $S_{i}$ as follows:
\begin{align}\label{eqFTLBOmutation}
S^{new}_{i,k} =  \begin{cases}
S^{old}_{i,k} + z,  &\text{ if }  w_0 < \mu \\
S^{old}_{i,k}&\text{ otherwise}  
\end{cases} 
\end{align}
where $k \in \{1,2,...,D\}$ is the $k$th dimension of $S_i$, $\mu$ is the mutation probability, and $z$ a mutation variable. $\mu$ is calculated as: 
\begin{equation}\label{eqFTLBOmu_prob}
\mu = 1 - \xi /\xi_{\text{max}},
\end{equation}
where $\xi$ and $\xi_{\text{max}}$ are respectively the fitness values corresponding to $A_i$ and the current best solution. $z$ is computed based on a chaotic sequence as follows: 
\begin{equation}\label{eqFTLBOmu}
z = 2 \times X_n - 1
\end{equation}
where $X_n$ represents the value of chaotic iteration $n$. This value is computed as:
\begin{equation}\label{eqFTLBOXn}
X_{n+1} = 4.0 \times X_n \times (1 - X_n),
\end{equation}
The initial value $X_0$ is randomly selected from the range $[0, 1]$. 

During the mutation process, the mutation variable is used to search for a better solution near the current best student. An elite strategy is then applied to replace the worst student $S_i$ as follows:
\begin{align}\label{eqFTLBOelite}
S_i =  \begin{cases} 
S^{new}_{i} &\text{ if } f(S^{new}_{i}) < f(S_i) \\
S_i &\text{ otherwise}
\end{cases}
\end{align}
where $f(\cdot)$ represents the objective function. 

After applying the mutation and elite selection strategies, we further enhance the solutions by utilizing a multi-subject approach described in \cite{rao2012elitist}. This approach improves the learning phase by allowing students to interact with each other within $m$ subjects. Let $i,k, \forall i \neq k$ be two random index values for two students, the learning phase in Eq. (\ref{eqTLBOlearner}) is updated for multi-subject as follows:

\begin{equation}\label{eqFTLBOlearner}
	S^{new}_{i,j} =  S^{old}_{i,j} + w_{i,j} \vert S_{i,j} - S_{k,j} \vert.
\end{equation}

\section{Algorithm Implementation} \label{secImplementation} 

\begin{figure}
	\centering
	\includegraphics[width=7cm]{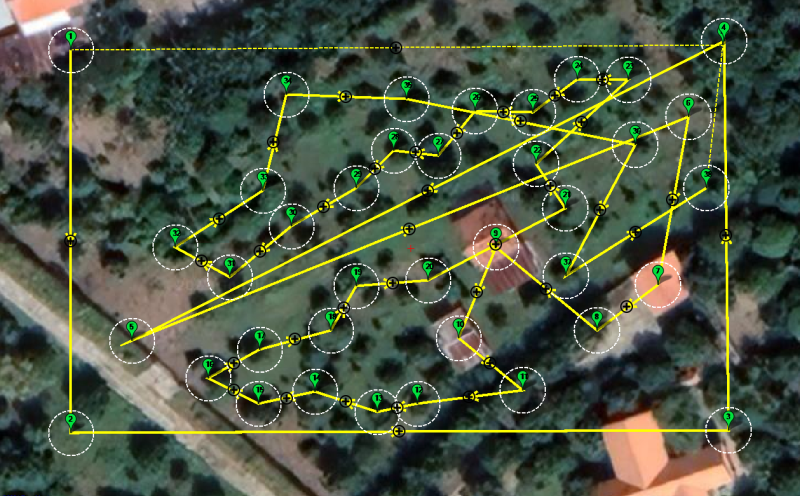}
	\caption{Working area with the obstacles identified}
	\label{Data_Acquisition}
	\vspace{-3mm}
\end{figure}

The implementation of the formation algorithm for a surveying task starts with setting parameters of the working area. They are determined by loading satellite images of the surveying area and selecting their upper ($UB$) and lower ($LB$) boundary points, as shown in Figure \ref{Data_Acquisition}. The maximum and minimum flight altitudes, $h_{max}$ and $h_{min}$,  are then set for the UAVs. Based on them, obstacles are identified and modeled by cylinders whose radii represent their size. The start and end locations, ($P_s$) and ($P_f$), are also selected. These setting parameters are used as input to the FTLBO, whose pseudo-code is shown in Figure \ref{figFTLBOpseudocode}. 

\begin{figure}[h!]
	{\fontsize{7}{6}\selectfont
	\removelatexerror
		\begin{algorithm}[H]
		\SetAlgoLined
			\tcc{Initialisation:}
			\quad Initialise the working environment by loading setting parameters\;
			\quad Initialize FTLBO parameters: $L$, $number\_of\_student$, $number\_of\_subject$, $number\_of\_decision\_variable$, $violation\_cost\_V$,...; \\
			
			\quad Create random paths from $P_s$ to $P_f$:\\
			\ForEach {$i < $\textit{(number\_of\_student)}} {
				\ForEach {$ j < $ \textit{(number\_of\_decision\_variable)}} {
					$P(i,j) = \text{rand} \times [UB_j - LB_j] + LB_j$; 
				}
			}
			\quad Initialize the chaos sequence:
			\quad $\xi = \xi + N$\\
			\quad $X =$ rand; \\

			\tcc{Path Planning:}
			\tcc{Teaching phase:}
			\While{$\xi < \xi_{\text{max}}$}{
    			Update chaos sequence value;
    			\tcc*[f]{using (\ref{eqFTLBOXn})}\\
    			\ForEach {$i < (number\_of\_decision\_variable)$} {
					\ForEach {$j < (number\_of\_decision\_variable)$} {
						\While{rand $< 1 – \xi/\xi_{\text{max}}$}{
							Update $S^{new}_{i,k}$\; 							\tcc*[f]{using (\ref{eqTLBOlearner})}
						}
					} 
				} 
				\quad Evaluate the fitness of $S^{new}_{i,k}$ \;
				\quad $\xi = \xi + 1$\;
				\While{$S^{new}_{i,k} > S_w$}{
							Update the worst student $S_w$ ;
							\tcc*[f]{using (\ref{eqFTLBOelite})}	
						}
    			\tcc{Learning phase:}
    			\ForEach {$S_m$ < $S_n$} {
					\ForEach {$j < (number\_of\_subject)$} {
						Select two random $S_m \neq S_n$\;
						Update the students' knowledge\; \tcc*[f]{using equation in (\ref{eqFTLBOlearner})}\\
					} 
				Evaluate the fitness value of $S^{new}$\;
				$\xi = \xi + N$;
				} 
			\quad Check $Violation\_cost\_V$; 			\tcc*[f]{using (\ref{eqViolation})}\\
			\quad Find $Best\_S$\;
			\quad Update $S$ and $V$; 
			\tcc*[f]{using (\ref{eqFTLBOelite}), (\ref{eqFTLBOlearner}), and (\ref{eqViolation})}
			}
		\quad return $Best\_S$.
		\end{algorithm}
	\caption{Pseudo code of the FTLBO for formation path planning.}
	\label{figFTLBOpseudocode}
  }
\end{figure}

\section{Experiments} \label{results}
A number of experiments have been conducted to evaluate the performance of the proposed algorithm with details as follows.
\subsection{Experimental setup}

In the experiment, the task is to survey an area of interest using a group of three UAVs, as shown in Figure \ref{Data_Acquisition}. A ground control station software, named Mission Planner, is used to gather initial information about the farm and its working conditions based on satellite images. The experimental area is a rectangular defined by two opposite vertices $\Gamma_1 = \{12.2335526,109.1144313\}$, $\Gamma_2 = \{112.2331044, 109.1152252\}$. The maximum and minimum altitudes are set to $h_{\text{max}} = 7$ m and $h_{\text{min}} = 2$ m, respectively. With these settings,  a number of obstacles are identified, each with a different radius as shown in Figure \ref{Data_Acquisition}. The start and end locations of the UAVs are set to $P_s = \{12.2332099, 109.1145051\}$ and $P_f = \{112.233474, 109.1151763\}$. The UAVs used are 3DR Solo quadcopters  \cite{Hoang2017AT}. Each UAV is fitted with necessary devices for task execution.

For the TLBO, the class size is chosen as $N= 100$, the paths have  10 waypoints, and $L = 150$ iterations. The initial positions of the UAVs with respect to the centroid of the formation chosen as follows: $\Delta P_1 = [0, 0, 2]$ m, $\Delta P_2 = [3, 0, -1]$ m, and $\Delta P_3 = [-3, 0, -1]$ m.
\subsection{Results}
\begin{figure}
	\centering
	\includegraphics[width=7cm,clip]{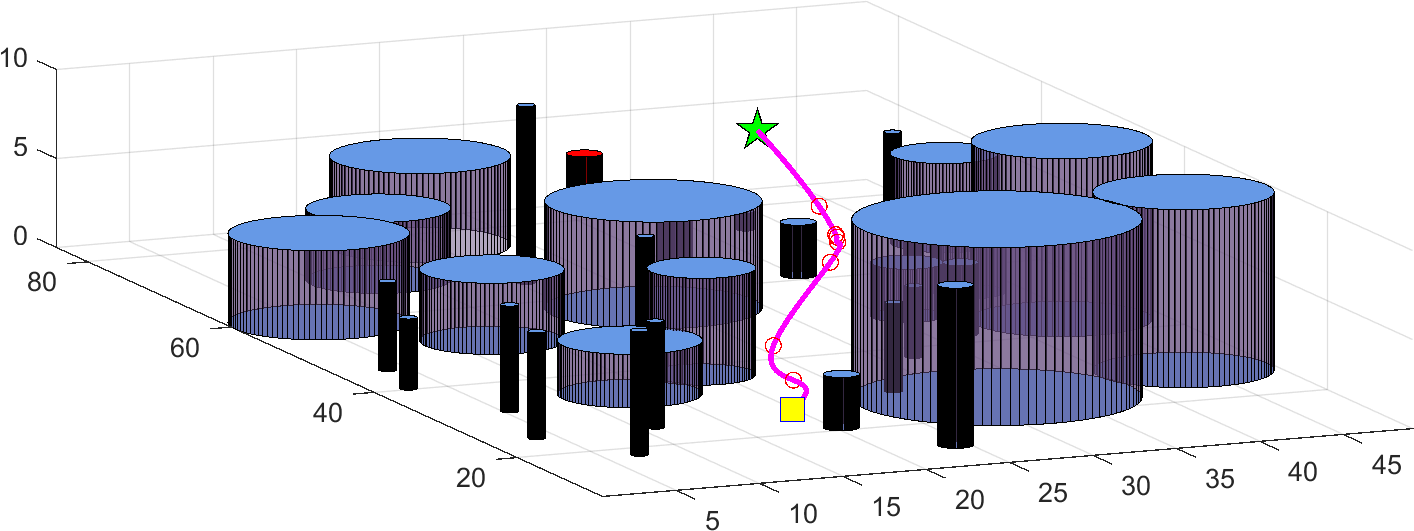}
	\caption{Generated path for the formation's centroid}
	\label{FormationPath}
	\vspace{-3mm}
\end{figure}

\begin{figure}
	\centering
	\includegraphics[width=7cm,clip]{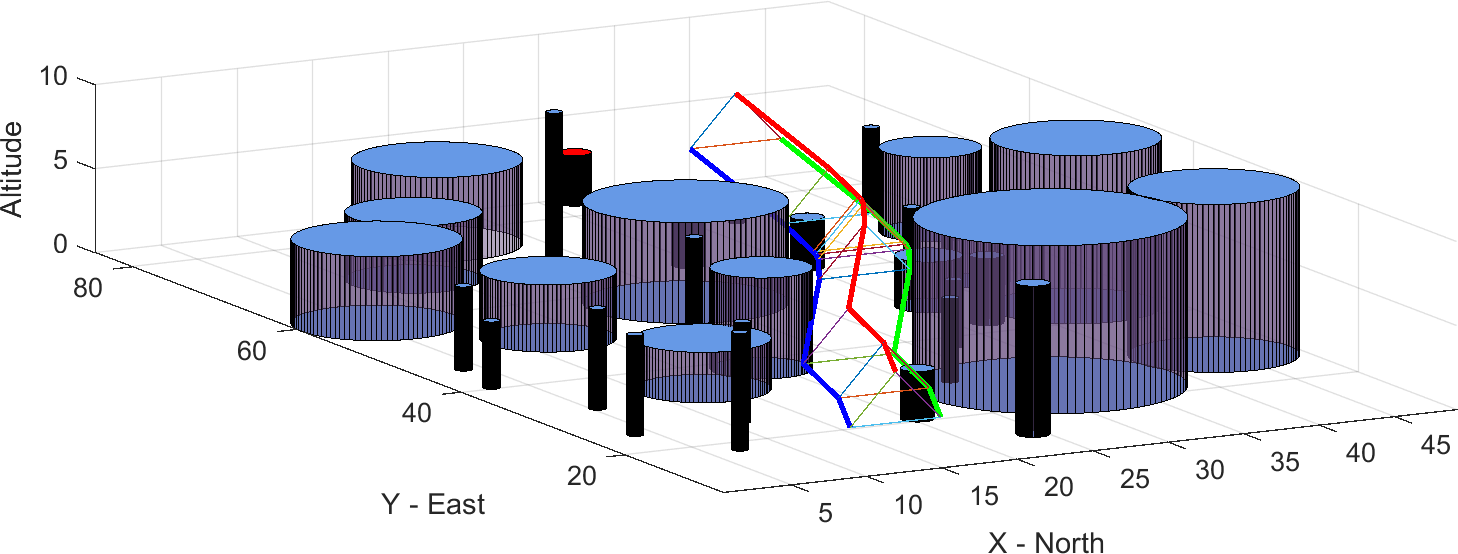}
	\caption{Generated paths for three UAVs}
	\label{UAVsPath}
	\vspace{-3mm}
\end{figure}

\begin{figure}
	\centering
	\includegraphics[width=6cm,clip]{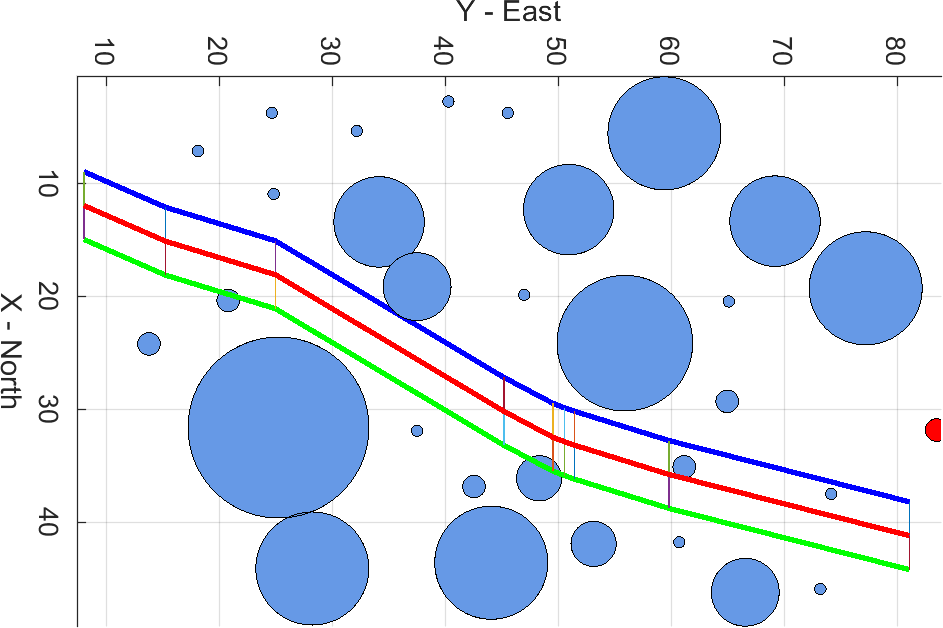}
	\caption{Projected paths on the ground (x-y plane)}
	\label{Ground_paths}
	\vspace{-3mm}
\end{figure}

\begin{figure}
	\centering
	\includegraphics[width=6cm,clip]{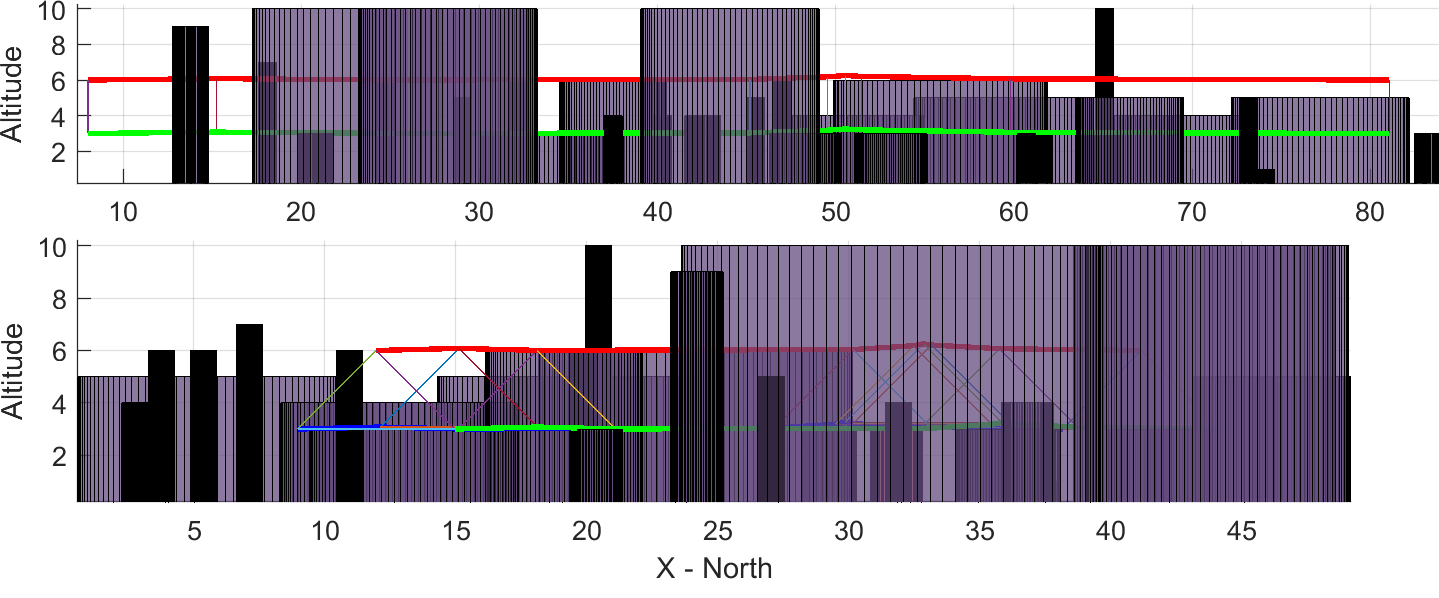}
	\caption{Projected paths on vertical (y-z and x-z) planes}
	\label{Vertical_paths}
	\vspace{-3mm}
\end{figure}

Figures \ref{FormationPath} and \ref{UAVsPath} show the 3D paths generated for the formation's centroid and three UAVs in a complex environment. It can be seen that the paths reach the goal location while avoiding obstacles. Figure \ref{UAVsPath} shows that the triangular shape of the UAVs is maintained at every waypoint. This formation can be further confirmed in Figures \ref{Ground_paths} and \ref{Vertical_paths} which show the paths projected onto the ground and vertical planes.

\begin{figure}
	\centering
	\includegraphics[width=8cm]{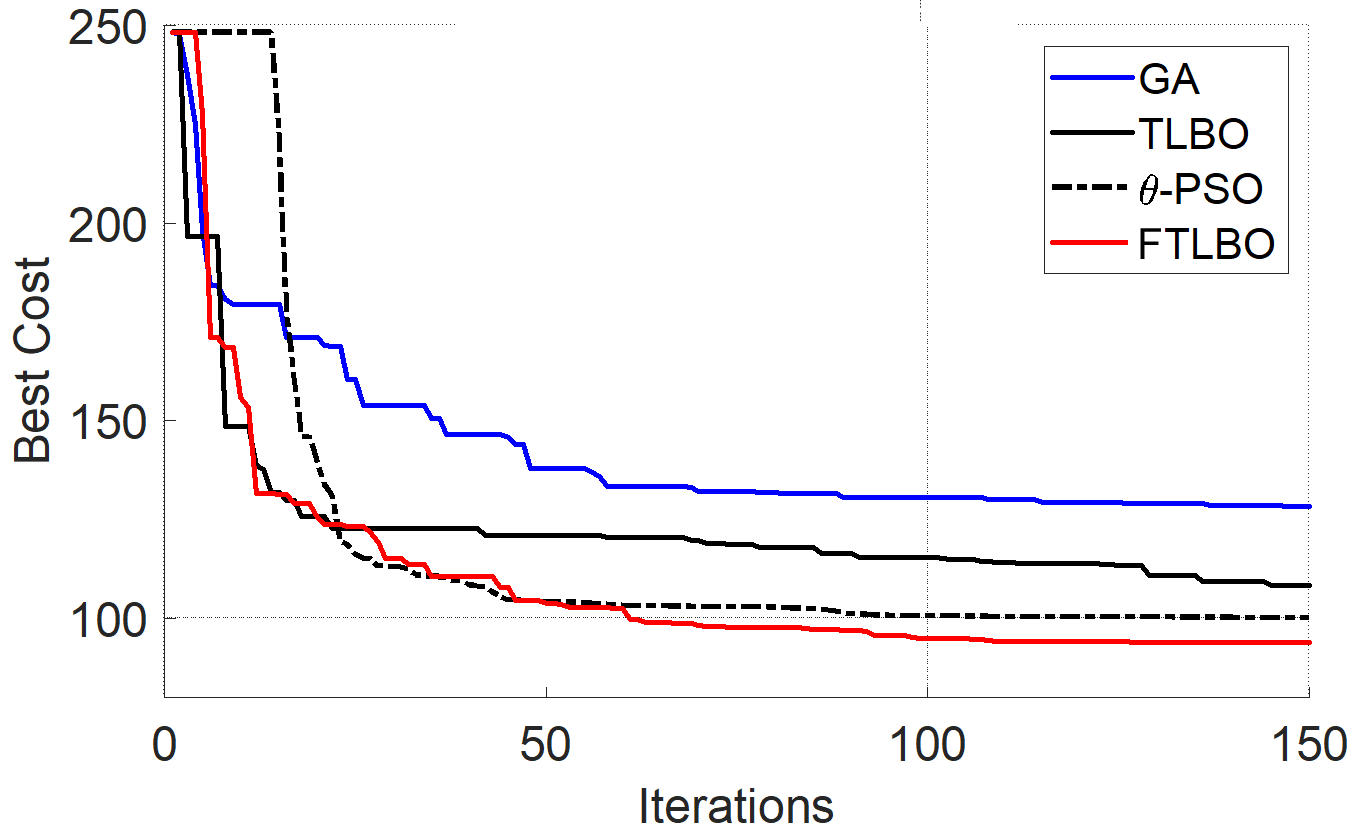}
	\caption{Convergence comparison}
	\label{Comparison}
	\vspace{-3mm}
\end{figure}

In another experiment, the performance of our proposed FTLBO is compared with state-of-the-art metaheuristic algorithms such as GA, TLBO, and $\theta$-PSO \cite{hoang2020}. Their convergence is shown in Figure \ref{Comparison}. It can be seen that FTLBO converges to the smallest fitness value implying that it can find best solutions. These results are further supported by the data in Table \ref{Table2}, which presents the converged cost values of the algorithms.
\begin{table}
\centering
	\caption{Performance comparison of path planning algorithms} \label{Table2} 
	\begin{tabular}{llll}
		\hline
		Algorithm	& Min cost 		& Max cost 	& Iterations \\ \hline
		GA	 		& 129.04	 		& 248.28		& 127\\
		TLBO 		& 108.31	 		& 248.28		& 146\\
		$\theta$-PSO& 100.63			& 248.28		& 96\\
		FTLBO 		& 94.70	 		& 248.28		& 107\\
		\hline 
	\end{tabular}
\end{table}

\subsection{Validation with real UAVs}
\begin{figure}
	\centering
	\includegraphics[width=8cm]{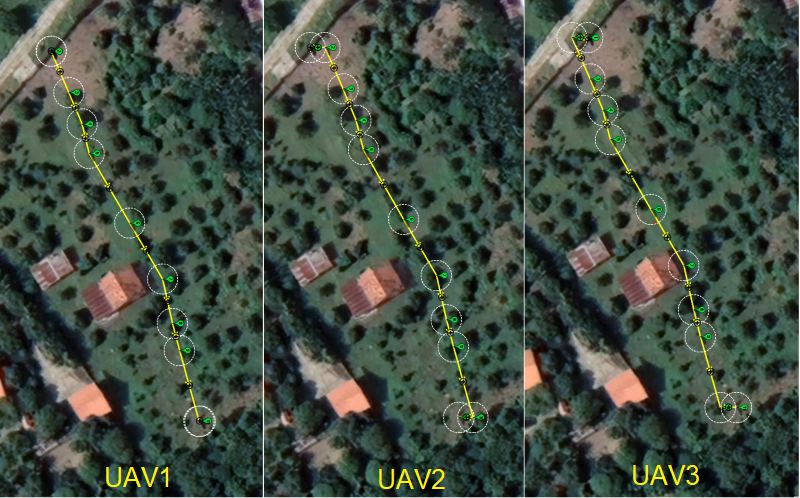}
	\caption{Trajectory commands for UAVs’ controllers }
	\label{Path_commands}
	\vspace{-3mm}
\end{figure}

To evaluate the validity of the paths generated, we have uploaded them to the UAVs via Mission Planner for autonomous flights as shown in Figure \ref{Path_commands}. The results show that the planned paths are smooth enough for the UAVs to follow. More importantly, the paths also navigate the UAVs to avoid obstacles in the environment, ensure safe operation of the UAVs. The results confirm the validity and practicality of the proposed algorithm.

\section{Conclusion}
This paper presents a novel approach to addressing the path planning problem for multi-UAV formations in orchard monitoring applications. An augmented teaching-learning-based optimization method is proposed to determine obstacle-free paths for the entire UAV group by minimizing a fitness function that considers shortest paths, safe operation, and effective mission execution. The generated paths ensure that individual UAVs maintain the formation shape relative to the group's centroid. The study also highlights the use of satellite maps to enhance real-world functionality. Experiments and comparisons confirm the feasibility of the proposed path planning algorithm. Future work will focus on applying this technique in practical scenarios.


%
%
%

%
%
\bibliography{IEEEabrv,bibi}
\end{document}